\documentclass[lettersize, journal]{IEEEtran}
\usepackage{amsmath,amsfonts}
\usepackage{algorithmic}
\usepackage{array}
\usepackage[caption=false,font=normalsize,labelfont=sf,textfont=sf]{subfig}
\usepackage{textcomp}
\usepackage{stfloats}
\usepackage{url}
\usepackage{verbatim}
\usepackage{graphicx}
\usepackage{amssymb}
\usepackage{amsmath}
\usepackage{multirow}
\usepackage{multicol}
\usepackage{booktabs} 
\usepackage{bbding} 
\def\BibTeX{{\rm B\kern-.05em{\sc i\kern-.025em b}\kern-.08em
    T\kern-.1667em\lower.7ex\hbox{E}\kern-.125emX}}
\usepackage{balance}
\usepackage[pdftex,hyperfigures,pagebackref=false]{hyperref}
\hypersetup{
    colorlinks=true,
    linkcolor=blue,
    citecolor=blue,
    urlcolor=blue,
    pdfstartview=FitH
}
\begin{document}
\title{Representation Space Constrained Learning with Modality Decoupling for Multimodal Object Detection}
\author{
       \IEEEauthorblockN{YiKang Shao, Tao Shi\IEEEauthorrefmark{1}}
\thanks{
\IEEEauthorrefmark{1} Corresponding author: Tao Shi.
Yikang Shao, Tao Shi are with the school of reliability and systems engineering, Beihang University, Beijing, 100191. China.
(email: shaoyikang@buaa.edu.cn).}}
\markboth{Journal of \LaTeX\ Class Files,~Vol.~18, No.~9, September~2020}%
{How to Use the IEEEtran \LaTeX \ Templates}
\maketitle

\begin{abstract} Multimodal object detection has attracted significant attention in both academia and industry for its enhanced robustness. Although numerous studies have focused on improving modality fusion strategies, most neglect fusion degradation, and none provide a theoretical analysis of its underlying causes. To fill this gap, this paper presents a systematic theoretical investigation of fusion degradation in multimodal detection and identifies two key optimization deficiencies: (1) the gradients of unimodal branch backbones are severely suppressed under multimodal architectures, resulting in under-optimization of the unimodal branches; (2) disparities in modality quality cause weaker modalities to experience stronger gradient suppression, which in turn results in imbalanced modality learning. To address these issues, this paper proposes a Representation Space Constrained Learning with Modality Decoupling (RSC-MD) method, which consists of two modules. The RSC module and the MD module are designed to respectively amplify the suppressed gradients and eliminate inter-modality coupling interference as well as modality imbalance, thereby enabling the comprehensive optimization of each modality-specific backbone. Extensive experiments conducted on the FLIR, LLVIP, M3FD, and MFAD datasets demonstrate that the proposed method effectively alleviates fusion degradation and achieves state-of-the-art performance across multiple benchmarks. The code and training procedures will be released at \url{https://github.com/yikangshao/RSC-MD}.
\end{abstract}


\section{Introduction}
\IEEEPARstart{M}{ultimodal} object detection\cite{ref1} has gained significant traction among researchers and engineers in recent years. By incorporating information from multiple modalities and establishing commonalities or complementarities between them, it achieves objectives unattainable through unimodal approaches or addresses novel challenges, garnering attention across diverse fields. Examples include object perception in autonomous driving\cite{ref2}, video surveillance and detection for urban security\cite{ref3}, and vehicle and object recognition in drone aerial imaging\cite{ref4}. 
Although significant progress has been made in unimodal object detection, relying solely on visible (VIS) light for detection still suffers from insufficient robustness and generalization under adverse imaging conditions such as weather changes\cite{ref5,ref6}. 
Infrared (IR) imaging is formed through thermal radiation, and its images are characterized by limited color diversity and low resolution. Using only IR images therefore greatly restricts the representational capacity of the model\cite{ref7},\cite{ref13}. 
Current studies combine visible and infrared images for object detection. By learning multimodal features, models can capture shared characteristics and fuse complementary semantic information, thus enhancing their representational capacity. For example, when visible images are degraded by adverse weather, infrared contours can supplement missing details, while clear textures in visible images can compensate for indistinct infrared features, leading to more robust and generalizable detection performance\cite{ref6, ref7}.

In recent years, this research area has witnessed remarkable progress\cite{ref8},\cite{ref9},\cite{ref10}. Existing approaches can be broadly categorized into two groups: two-stage and one-stage methods. Two-stage methods first perform image-level fusion of multiple modalities\cite{ref14},\cite{ref15},\cite{ref16} and subsequently conduct object detection on the fused images. However, these methods suffer from an inherent limitation—independent optimization of fusion and detection leads to semantic misalignment and potential inconsistency between feature representations.

Researchers have proposed one-stage frameworks that integrate image fusion and object detection into a unified learning process. Recent studies\cite{ref17},\cite{ref18},\cite{ref19} have shown that feature-level multimodal fusion achieves superior performance compared with image-level or decision-level fusion.
One-stage methods mainly focus on improving multimodal fusion strategies to enhance detection performance\cite{ref11},\cite{ref12},\cite{ref13} or employ sophisticated attention mechanisms to better exploit cross-modal complementarity\cite{ref7},\cite{ref20},\cite{ref21}. Although these approaches have achieved notable results, they overlook a fundamental yet counterintuitive issue—certain objects that can be accurately detected by unimodal detectors fail to be detected by multimodal ones. The study\cite{ref5} investigated this phenomenon and attributed it to insufficient learning within unimodal branches during multimodal training, referring to it as the fusion degradation phenomenon.

Although study\cite{ref5} attempted to mitigate this issue by introducing additional knowledge distillation\cite{ref22} to strengthen unimodal learning, it did not fundamentally reveal the underlying cause of unimodal learning insufficiency, nor did it provide a theoretical explanation for how unimodal degradation introduces optimization defects in multimodal models. 
Some studies have also addressed similar issues of multimodal optimization defects. Study\cite{ref23} suggested that large variations in image acquisition conditions or technical challenges leading to modality degradation can result in extreme modality imbalance, thereby impairing model performance. Studies\cite{ref24,ref25}  proposed that incompatible information across modalities may cause fusion conflicts, which constrain model optimization. However, these analyses focus only on modality fusion or data quality and still do not identify the fundamental architectural causes of multimodal optimization defects.

To address this issue, this study provides a theoretical analysis of fusion degradation in multimodal detection, demonstrating that current multimodal object detection architectures exhibit optimization defects. Even well-performing single-modality branches are affected and cannot achieve the performance of models trained on single-modality data.

Specifically, this study theoretically demonstrates two optimization defects in multimodal object detection frameworks.\textbf{ First}, the gradients of unimodal branch backbones are excessively suppressed by the fusion module, resulting in under-optimization of the unimodal backbones.\textbf{ Second}, due to varying quality among modalities, this gradient suppression exhibits an amplification effect across modalities: weaker modalities experience stronger gradient suppression, causing the model to overly rely on stronger modalities while neglecting weaker ones, which leads to optimization imbalance among the unimodal branches.

To remedy the optimization deficiencies identified through theoretical analysis, this paper proposes a Representation Space Constrained Learning with Modality Decoupling (RSC-MD) method. The proposed method consists of two major components: the Representation Space Constraint (RSC) module and the Modality Decoupling (MD) module.

Specifically, the RSC module imposes an auxiliary representational constraint on each backbone network to amplify the gradients suppressed by the fusion module, thereby promoting sufficient learning within unimodal backbones. Furthermore, the MD module aims to eliminate cross-modal competitive learning and interference induced by modality coupling. By decoupling the backbone networks of different modalities and enabling their independent optimization, the MD module prevents the optimization deficiencies that arise from gradient suppression and inter-modality competition between strong and weak modalities, thereby mitigating the modality imbalance problem during multimodal learning.

In summary, the contributions of this paper are as follows:
\begin{list}{}{}
\item{ $\bullet$ \textbf{This paper theoretically demonstrates the existence of unimodal under-optimization in multimodal object detection:} the multimodal fusion module hinders the optimization of each modality-specific backbone network, resulting in under-optimization of the unimodal branches.}
\item{ $\bullet$ \textbf{This paper theoretically demonstrates the imbalanced optimization defect in multimodal object detection}: weaker modalities experience greater optimization suppression, causing the model to prioritize dominant modalities while neglecting weaker ones, thereby failing to effectively leverage the complementary advantages of multimodality.}
\item {$\bullet$ This paper proposes a Representation Space Constraint (RSC) module for unimodal backbone networks. By imposing auxiliary representational learning constraints on each modality backbone, the module amplifies suppressed gradients and promotes sufficient learning within unimodal branches.}
\item{ $\bullet$ This paper proposes a Modality Decoupling (MD) module for multimodal detection. By employing a modality decoupling strategy, the MD module enables independent optimization of each modality, thereby eliminating inter-modal conflicts.}
\end{list}

\section{Related Work }
\subsection{Multimodal Object Detection}
In recent years, numerous studies in the field of multimodal object detection have achieved remarkable success. These advancements have been successfully applied across multiple domains, including autonomous driving, robotics engineering, and satellite remote sensing imagery\cite{ref29},\cite{ref35},\cite{ref36},\cite{ref37}, thereby attracting increasing attention to multimodal object detection. According to prevailing taxonomies, multimodal object detection methods are primarily categorized into three classes based on fusion stages: early fusion, intermediate fusion, and late fusion—also referred to as pixel-level, feature-level, and decision-level fusion, respectively. Certain early fusion approaches are termed two-stage detection methods in multimodal object detection. These methods first fuse multimodal images and then perform object detection on the fused result\cite{ref14},\cite{ref15},\cite{ref16},\cite{ref38,ref39,ref40,ref41,ref42}. However, pixel-level fusion generally incurs high computational costs and large model sizes, while often failing to achieve satisfactory performance and inference speed\cite{ref31}. Moreover, two-stage methods are widely recognized to suffer from optimization conflicts caused by the decoupling of fusion and detection processes\cite{ref13}.

Late-stage or decision-level fusion methods\cite{ref43} aim to perform detection using independent detectors for each modality and to combine their outputs in order to enhance the robustness of the final results. However, decision-level fusion is constrained by conflicts and imbalanced dependencies among the independent detectors, resulting in performance inferior to that of early and middle fusion approaches.

An increasing number of studies have demonstrated that feature-level fusion methods generally outperform the other two paradigms and have been extensively employed in multimodal object detection research\cite{ref8},\cite{ref44, ref45}. In this study\cite{ref8}, a Transformer-based architecture was adopted to achieve mid-level feature fusion, and in a subsequent study\cite{ref45}, a cross-modal attention fusion module was introduced to enhance modality fusion and thereby improve detection performance. Reference\cite{ref46} proposed an uncertainty-aware fusion approach to address calibration errors and modality discrepancies within paired images.
With the widespread adoption of Transformer architectures, attention-based fusion methods have garnered considerable research interest. By constructing sophisticated attention mechanisms, these approaches aim to optimize the fusion of visible and infrared modalities, thereby further improving detection accuracy\cite{ref47,ref48,ref49,ref50,ref51,ref52,ref58}. Similar to Transformer-based designs, Mamba-based modality fusion frameworks\cite{ref26},\cite{ref37},\cite{ref53} have also achieved promising results in multimodal detection. Furthermore, the successful application of knowledge distillation\cite{ref22} in object detection has inspired subsequent research efforts. Distillation-based frameworks that guide networks to more effectively extract modality-specific representations and facilitate cross-modal feature fusion have become one of the prevailing research directions in this field\cite{ref5},\cite{ref54},\cite{ref55}.

However, these studies primarily focus on exploring more effective modality fusion strategies to enhance the performance of multimodal detection models, while overlooking the phenomenon of fusion degradation inherent in current feature-level fusion methods and the underlying architectural deficiencies behind it.

\subsection{Modality Conflict and Modality Imbalance in Multimodal Detection}
Recent studies have begun to address the issues of modality imbalance and fusion degradation in multimodal detection methods\cite{ref5},\cite{ref34}. However, most studies still attribute this problem to fusion conflicts caused by discrepancies among modalities\cite{ref37}. Reference\cite{ref56} points out that most existing studies rely on integrating complementary information from different modalities while overlooking the semantic conflicts induced by their inherent discrepancies. To mitigate this issue, it introduces a modality conflict correction approach. The work in\cite{ref24} argues that indistinguishable intra-modal features can cause single-modality interference and weaken the dominant modality's representation. To address this, a confidence-based strategy is proposed to eliminate such interference. The study in\cite{ref25} attributes inter-modal heterogeneity to differences in task-related information content within each modality, suggesting that a significant imbalance exists in the amount of information each modality carries. It proposes a dynamic modality information balancing method to alleviate this issue. According to\cite{ref57}, feature-level fusion methods inherently suffer from modality imbalance, and dynamic dropout with threshold masking is introduced to compensate for this problem. Reference\cite{ref58} also identifies imbalance in multimodal detection and proposes the use of indicative illumination signals to guide the attention computation for mitigation.

Recent studies have identified differences in modality information content and feature representations across modalities, attributing them to the causes of fusion conflicts and imbalanced modality learning. These studies primarily focus on improving fusion strategies to achieve more effective modality integration, thereby alleviating conflicts and imbalance during multimodal learning. Although such efforts have indeed enhanced overall model performance, they have largely overlooked the intrinsic architectural deficiencies of existing detection frameworks. References\cite{ref5, ref30} employ knowledge distillation techniques to strengthen modality-specific feature extraction and mitigate the under-optimization of weaker modalities. However, they fail to provide a theoretical explanation for the underlying cause of such under-optimization. Similarly, the works in\cite{ref27},\cite{ref28},\cite{ref31} also recognize the presence of modality imbalance, but their approaches essentially address the issue by handling data noise or by adaptively adjusting fusion mechanisms according to modality contributions.
Studies\cite{ref32},\cite{ref33},\cite{ref34} build on imbalance-related research in the multimodal domain and reveal that an intrinsic disparity exists between strong and weak modalities within multimodal architectures. However, they fail to provide theoretical evidence from the perspective of the multimodal detection architecture itself.

In summary, recent studies concerning modality conflicts and imbalance have predominantly focused on refining fusion mechanisms, while few have investigated the inherent deficiencies within the architectural design of multimodal detection frameworks.
\begin{figure*}[!t]
\centering
\includegraphics[width=7in]{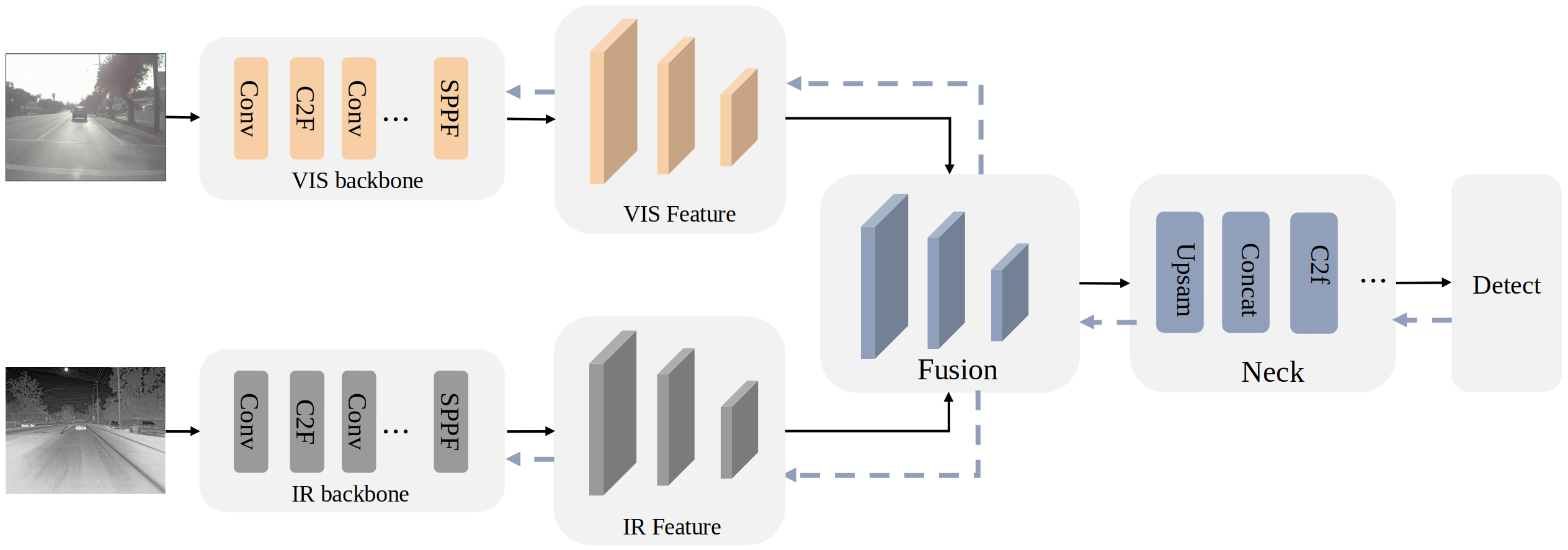}
\caption{Architectural Diagram of a Dual-Modal Object Detection Framework Using Naive Addition for Feature Fusion.}
\label{naiveAdd}
\end{figure*}

\section{Method}
In this section, this paper first provides a theoretical analysis to demonstrate two optimization deficiencies inherent in current multimodal object detection frameworks and subsequently proposes a Representation Space Constrained Learning with Modality Decoupling (RSC-MD) approach to address these deficiencies.
\subsection{Theoretical Analysis of Defects in Modality Optimization for Multimodal Object Detection}
To address defects such as fusion degradation in multimodal detection, this study employs theoretical analysis to elucidate both the origin of these deficiencies and their impact on the overall model. For a given sample $x_i$, a multimodal object detection model accepts inputs from two modalities, $m_1$ and $m_2$, such that the sample can be represented as $x_i= (x^{m_1}_i, x^{m_2}_i)$. In alignment with the current state-of-the-art object detection model YOLO, this work utilizes different layers of the feature pyramid as detection features, commonly including the outputs of layers $P3$, $P4$, $P5$, which are collectively abstracted here as $B_i$. Similarly, the structure of the feature extraction network is abstracted uniformly; for the input modality $x^{m_1}_i$, the resulting feature representation can be expressed as:

\begin{equation}
\label{deqn_ex1}
f^{m_1}_i = B_1(x^{m_1}_i\ ;\  \theta_1)
\end{equation}
where, $f^{m_1}_1$ represents the image features of modality $m_1$ extracted by the feature extraction layer $backbone_1$, and $\theta_1$ denotes the parameters of $B_1$, where the outputs of layers $P3$, $P4$, and $P5$ are included within $B_1$.

Similarly, the modal representation of $x^{m_2}_i$ can be obtained:
\begin{equation}
\label{deqn_ex2}
f^{m_2}_i = B_2(x^{m_2}_i\ ;\  \theta_2)
\end{equation}

Consistent with the majority of existing studies, this work employs feature-level intermediate fusion as the modality feature fusion method. Accordingly, the fusion module can be abstractly represented by the following formulation:
\begin{equation}
\label{deqn_ex3}
z_i  = W^{m_1}_k f^{m_1}_{i} + W^{m_2}_k f^{m_2}_{i}
\end{equation}
Here, $f^{m_1}_1$ and $f^{m_2}_2$  denote the outputs of the aforementioned feature extraction networks, which serve as the inputs to the fusion module. Similar to the backbone networks, the detection network module is abstracted as a complex composite function comprising multiple hierarchical levels. Although different versions and architectures of object detection algorithms may employ distinct computational procedures, its forward computation can be abstractly expressed, in a general sense, as follows:
\begin{equation}
\label{deqn_ex4}
C_i= \Phi(z_i  ; \ \theta_{\phi})
\end{equation}
where, $\theta_\phi$ denotes the parameters of the detection module function.

For the loss function, considering the currently widely adopted YOLO model, its loss function can be expressed as follows:
\begin{equation}
\label{deqn_ex5}
L=\lambda_{box}L_{box}+\lambda_{cls}L_{cls}+\lambda_{dfl}L_{dfl}
\end{equation}
Among these, the classification loss is expressed as:
\begin{equation}
\label{deqn_ex6}
L_{cls} = -\frac{1}{n} \sum_{i=0} ^{n}[y_ilog\sigma(\hat{y_i}) + (1-y_i)log(1-\sigma(\hat{y_i}))]
\end{equation}
According to the chain rule of gradients, the gradient of the classification loss in multimodal object detection backpropagated to $backbone_1$ can be expressed as follows:
\begin{equation}
\label{deqn_ex7}
g_{B_1} =  \frac{\partial L_{cls}}{\partial f(z)}\frac{\partial f(z)}{\partial z}\frac{\partial z}{\partial f_1}
\end{equation}
According to the above formula, the gradient propagated backward to $backbone_1$ can be calculated as:
\begin{equation}
\begin{split}  
\label{deqn_ex8}
g_{B_1} =& (\frac{1 }  {  1 + e^ {-(W_{k}^{m_1} f_i^{m_1} + b_1) - (W_{k}^{m_2} f_i^{m_2} + b_2)} } - 1) \theta_{\phi} \ \ \\ & \ \ ,  \ \ positive
\end{split}
\end{equation}
\begin{equation}
\begin{split}       
\label{deqn_ex9}
 g_{B_1} =  & (\frac{1 }  {  1 + e^ {-(W_{k}^{m_1} f_i^{m_1} + b_1) - (W_{k}^{m_2} f_i^{m_2} + b_2)} })\theta_{\phi}  \ \ \\ & \ \ ,  \ \ negative
\end{split}
\end{equation}
Here, $W_{k}^{m_1}$ and $W_{k}^{m_2}$ represent the corresponding parameters in the fusion module at an abstract mathematical level. The terms positive and negative denote the cases of positive and negative samples, respectively. Correspondingly, the gradient for the unimodal model, denoted as $g_{Uni}$, can be expressed as follows:
\begin{equation}
\label{deqn_ex10}
g_{Uni} = (\frac{1 }  {  1 + e^ {-(W_{k}^{m_1} f_i^{m_1} + b_1) }} - 1) \theta_{\phi}
  \ \  , \ \  \ \ positive
\end{equation}
\begin{equation}
\label{deqn_ex11}
g_{Uni} = (\frac{1 }  {  1 + e^ {-(W_{k}^{m_1} f_i^{m_1} + b_1) }}) \theta_{\phi}, \ \  \ \ negative
\end{equation}
Since the model employs the $SiLU$ activation function, the output feature values can be approximately regarded as non-negative. Its formulation is given as follows:
 
\begin{equation}
\label{deqn_ex12}
SiLU( x ) = x *\frac{1 }  {1 + e^{-x}}
\end{equation}
It can be inferred that the range of the $logits$ values after passing through the activation function is given by:
\begin{equation}
\label{deqn_ex13}
\begin{cases}
 \ \  \ \ SiLU(x) > 0,        \ \ \ \ \  \ \ \       x  > 0 \\[2ex]
 \ \  \ \ SiLU(x) \approx 0,        \ \ \ \ \  \ \   x  \leq 0
\end{cases}
\end{equation}
Based on the above results, it can be approximately assumed that the value of $e^{-(W_{k}^{m_2} f_i^{m_2})}$ is less than or equal to 1. By applying this conclusion to the gradient computations propagated back to the backbone in both multimodal and unimodal architectures, the following expressions can be obtained: 
\begin{equation}
\label{deqn_ex14}
e^ {-(W_{k}^{m_1} f_i^{m_1} + b)} \geq e^ {-(W_{k}^{m_1} f_i^{m_1} + W_{k}^{m_2} f_i^{m_2} + b)}
\end{equation}
Based on the foregoing analysis, the representation for positive samples can be derived as follows:
\begin{equation}
\label{deqn_ex15}
\begin{split}
& \frac{1}  { 1 + e^ {-(W_{k}^{m_1} f_i^{m_1} + b) } } - 1  < \\ & \frac{1}  {1 + e^ {-(W_{k}^{m_1} f_i^{m_1} + b) - (W_{k}^{m_2} f_i^{m_2})} }  - 1 < 0
\end{split}
\end{equation}

Due to the substantial suppression of gradients in the multimodal backbone networks compared with the unimodal case, the performance of the multimodal detection backbones converges more slowly and less effectively than that of unimodal networks, which severely limits the optimization of the model. Consequently, we identify the first optimization deficiency.

\textbf{Optimization Deficiency (1)}: In the multimodal architecture, the gradients propagated from the fusion detection module back to the backbone are significantly smaller than those in the unimodal case, resulting in under-optimization of the unimodal branches within the multimodal detection framework.

Due to differences in modality quality, the ease of learning varies across modalities. A widely acknowledged consensus in multimodal learning is that the model tends to prioritize the easier-to-learn modalities, falling into a predicament in which it focuses on the dominant modalities while neglecting the weaker ones, thereby substantially limiting overall model performance. In extreme cases, the performance of a multimodal detection model may even be inferior to that of a unimodal model, which contradicts the original intention of multimodal learning to effectively exploit information from multiple modalities and leverage complementary features.

Further analysis reveals that, under the current multimodal architecture, the gradients applied by the fusion detection module to the backbone are identical across modalities, which leads to imbalance in the optimization of the two modalities. Compared with the unimodal architecture, the gradients propagated to the backbone in multimodal detection include additional modality-specific terms, the magnitude of which can be expressed as follows:
\begin{equation}
\label{deqn_ex16}
e^ {-(W_{k}^{m_2} f_i^{m_2}) } < \ e^ {-(W_{k}^{m_1} f_i^{m_1}) }   , m_1 =weak Modality
\end{equation}
\begin{equation}
\label{deqn_ex17}
 e^ {-(W_{k}^{m_2} f_i^{m_2}) }  > \ e^ {-(W_{k}^{m_1} f_i^{m_1}) }   , m_2 = weak Modality
\end{equation}
where, $m_i = weak modality$ denotes that modality $m_i$ represents the weaker modality, which is more difficult to learn. In contrast, the other modality serves as the dominant modality, being relatively easier to learn and represent. If $m_2$ corresponds to the easier-to-learn modality, the relationship $W_{k}^{m_1} f_i^{m_1} < W_{k}^{m_2} f_i^{m_2}$ generally holds during representation learning. This is because the dominant modality converges faster and more effectively during optimization, causing its feature weight vector to be closer to the class center, thereby resulting in a larger inner product value\cite{ref59, ref60}.

By integrating the above formulations, the expression for the gradient difference between the multimodal and unimodal cases can be derived as follows:
\begin{equation}
\label{deqn_ex18} 
\begin{split}
&\frac{1}  {1 + e^ {-(W_{k}^{m_1} f_i^{m_1} + b) - (W_{k}^{m_2} f_i^{m_2})} }  - \frac{1}  { 1 + e^ {-(W_{k}^{m_1} f_i^{m_1} + b) } } 
>  
 \\ &  \frac{1}  {1 + e^ {-(W_{k}^{m_2} f_i^{m_2} + b) - (W_{k}^{m_1} f_i^{m_1})} } - \frac{1}  { 1 + e^ {-(W_{k}^{m_2} f_i^{m_2} + b) } } 
\\ & , \ \ m_1 =\ weak\  Modality 
\end{split}
\end{equation}
\begin{equation}
\label{deqn_ex19} 
\begin{split}
&\frac{1}  {1 + e^ {-(W_{k}^{m_1} f_i^{m_1} + b) - (W_{k}^{m_2} f_i^{m_2})} }  - \frac{1}  { 1 + e^ {-(W_{k}^{m_1} f_i^{m_1} + b) } } 
<  
 \\ &  \frac{1}  {1 + e^ {-(W_{k}^{m_2} f_i^{m_2} + b) - (W_{k}^{m_1} f_i^{m_1})} } - \frac{1}  { 1 + e^ {-(W_{k}^{m_2} f_i^{m_2} + b) } } 
\\ & , \ \ m_2 =\ weak\  Modality 
\end{split}
\end{equation}

Based on the above theoretical analyses, it can be concluded that the weak modality suffers from more pronounced gradient suppression compared with the strong modality, resulting in insufficient optimization of its feature extraction capability.

\textbf{Optimization Deficiency (2)}: The weak modality undergoes greater gradient suppression than the strong modality, resulting in imbalanced learning across unimodal branches. Consequently, the model tends to prioritize the dominant modality at the expense of the weaker one, thereby failing to effectively exploit the complementary advantages of multimodal information.

The same optimization deficiencies are present for negative samples. Recent studies in object detection\cite{ref63}, have demonstrated that during the optimization of detection models, there exists an imbalance between positive and negative samples. Existing optimization methods primarily aim to emphasize positive samples, treating them as the dominant contributors. However, according to the aforementioned theoretical analysis, the gradients applied by multimodal detection models for negative samples behave in a manner entirely opposite to that for positive samples. Specifically, in multimodal object detection, negative samples contribute larger gradients compared to unimodal models. This observation contradicts the principle of weighted positive and negative samples and instead imposes additional detrimental effects on model optimization. 
\begin{figure*}[!t]
\centering
\includegraphics[width=7.1in]{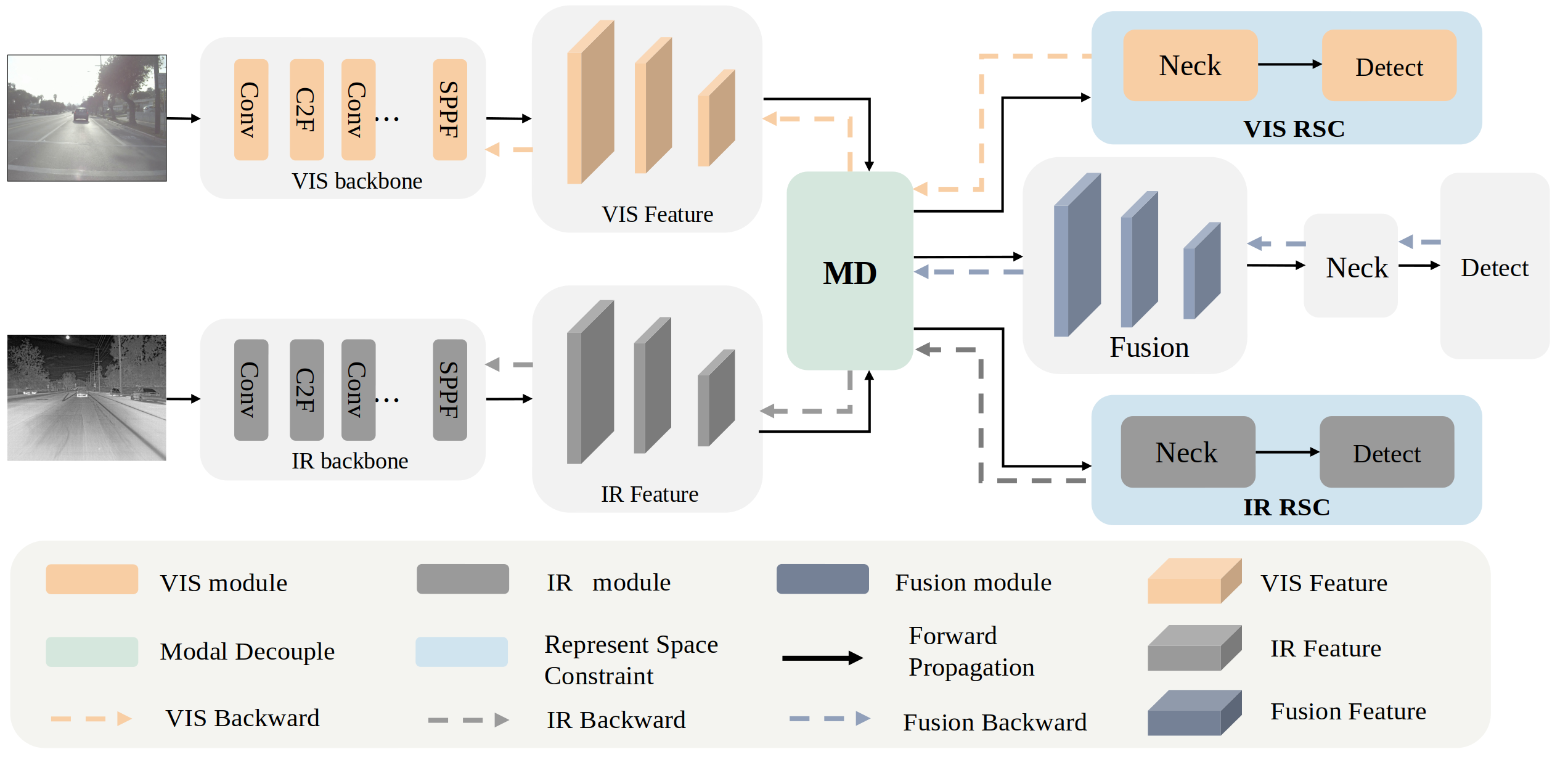}

\caption{Architecture Diagram of Representation Space Constrained Learning with Modality Decoupling Framework.}
\label{rsc-md}
\end{figure*}
Therefore, the conclusions drawn for positive samples also hold for negative samples: the increased negative sample gradients similarly impair the proper optimization of the model.

In summary, the above theoretical analysis clearly identifies the origins of fusion degradation in multimodal detection: optimization conflicts exist inherently among modules in multimodal object detection methods. The gradients propagated to the backbone networks of multimodal detection models are smaller than the corresponding unimodal gradients, and modalities of differing learning difficulty are subject to varying degrees of gradient suppression. This results in imbalanced learning across multiple modalities, further constraining the overall optimization of the detection task.

\subsection{Representation Space Constrained Learning with  Modality Decoupling}
By comparing the mathematical expressions derived from the theoretical analysis, it can be observed that the interference causing the optimization deficiencies originates from the other modality. This interference does not exist in unimodal learning but persists in multimodal learning due to improper operations involving logarithmic and exponential transformations, which prevent its correct elimination. Specifically, the term $W_{k}^{m_2} f_i^{m_2}$ from the other modality is not properly canceled during the derivative computation of $e^ {-(W_{k}^{m_2} f_i^{m_2}) }$ resulting in the two modalities being coupled during training and interfering with the optimization of their respective backbone networks.

Motivated by this observation, this study proposes an architectural innovation that decouples the coupled modalities, allowing each modality to learn representations independently without mutual interference. This study proposes a Representation Space Constrained Learning with Modality Decoupling (RSC-MD) method, as illustrated in Figure \ref{rsc-md}, which consists of two main components. The first component enhances the feature extraction capability of each unimodal branch backbone by imposing representational learning constraints within the multimodal architecture, wherein each backbone is equipped with an independent detection head and corresponding loss supervision. The second component mitigates optimization deficiencies caused by inter-modality coupling interference through modality decoupling, ensuring that the optimization of each unimodal backbone proceeds independently without interference from other modalities, while enabling the fusion module to effectively integrate multimodal features for joint optimization.

\subsubsection{Representation Space Constrained Learning}
To address the under-optimization of unimodal branches in multimodal detection models, this work introduces a Representation-Constrained Supervision (RCS) module, which imposes representation learning constraints on each unimodal backbone. RCS is designed to amplify the backpropagated gradients to strengthen the feature extraction capability of each modality and to ensure that the learned representations remain aligned with the original optimization direction of the corresponding unimodal branch. Specifically, RCS employs two additional detection heads that respectively receive the multi-scale feature maps produced by the backbones of the two modalities. The process can be formulated as follows:
\begin{equation}
\label{deqn_ex20}
\begin{cases}
Aux^{m_1} = H(f^{m_1}_i; \theta_{a_1}) \\[2ex]

Aux^{m_2} = H(f^{m_2}_i; \theta_{a_2})
\end{cases}
\end{equation}
where, $Aux^{m_1} $ and $Aux^{m_2} $ denote the outputs of auxiliary detection heads, while $H$ represents an abstract detection head module—a multi-level composite function. $\theta_{a_1}$ and $\theta_{a_2}$ denote their internal parameters. 
After the auxiliary heads $Aux^{m_1}$ and $Aux^{m_2}$ independently complete their forward computations, the backpropagated gradients are transmitted solely to the corresponding modality-specific backbones. In this way, each backbone receives a targeted representation constraint, guiding its feature space to progressively approach the representation space obtained under unimodal training. 

Accordingly, two auxiliary loss functions are added for the two auxiliary detection heads, expressed as:

\begin{equation}
\label{deqn_ex21}
\begin{cases}
L_{A_1} = Loss(Aux^{m_1} , Y) \\[2ex]

L_{A_2} = Loss(Aux^{m_2} , Y)
\end{cases}
\end{equation}
where, $Loss(\bullet)$ denotes the model loss computation. For consistency, it follows exactly the same formulation as the loss function used in the unimodal architecture.

The total model loss is expressed as:
\begin{equation}
\label{deqn_ex22}
L_{total} = \alpha L_{fusion} + \beta  L_{A_1} + \gamma L_{A_2}
\end{equation}
where, $\alpha, \beta, \gamma$ denote the constraint coefficients applied to different modalities, and they serve as hyperparameters of the entire model.

The RCS module imposes directional constraints on each unimodal branch backbone through auxiliary detection heads, guiding each backbone network to approach the representation space corresponding to its respective unimodal training. Consistent with prior studies such as\cite{ref5}, linear probing evaluations are conducted on the backbone networks of each modality. The results, shown in Figure \ref{rgb_perf} and Figure \ref{ir_perf}, indicate that although the auxiliary detection heads increase the previously suppressed gradients and improve overall model performance, the optimization of each unimodal branch backbone is still influenced by the presence of the other modality. Consequently, some unimodal branch backbones within the multimodal architecture fail to reach the performance levels obtained under unimodal training. This indicates that, although the gradient suppression deficiency is partially alleviated, the optimization deficiencies caused by the imbalance between strong and weak modalities, as well as the excessive negative-sample gradients, remain unaddressed.

To overcome this limitation, a modality decoupling method is proposed to eliminate these optimization deficiencies, ensuring that the backbone networks used for feature extraction in multimodal object detection are fully optimized and capable of providing more effective features for multimodal feature fusion.

\subsubsection{Modality-Decoupled Learning}
The purpose of the MD module is to eliminate the residual terms ${e^ {-(W_{k}^{m_2} f_i^{m_2}) } }$ or $e^ {-(W_{k}^{m_1} f_i^{m_1})}$ in the polynomial $ \frac{1}{1 + e^{-(W_{k}^{m_1} f_i^{m_1} + b) - (W_{k}^ {m_2} f_i^{m_2})} }$ which, due to modality coupling, are not correctly eliminated during backbone optimization in multimodal learning. In the aforementioned analysis, even after introducing additional representation constraints, optimization interference between modalities still persists. 
Although auxiliary detection heads can partially mitigate the gradient suppression on positive samples, the optimization deficiencies arising from imbalanced modality learning and excessive negative-sample gradients that undermine the dominance of positive samples remain unresolved.
At this stage, the backbone networks of different modalities are jointly constrained by the detection head of the fusion module and the newly introduced auxiliary detection heads. As a result, the model suffers from both imbalanced modality learning and interference from negative samples that weakens the dominance of positive samples, causing each unimodal backbone to remain unable to match or surpass the performance of single-modality models.

Building upon the RSC module, this study proposes a modality decoupling (MD) method, which is applied at the interface between the modality backbone networks and the fusion network to correctly map or directly discard backpropagated gradients. As illustrated in Figure \ref{rsc-md}, the MD module enables decoupled representation learning for each backbone network, thereby eliminating inter-modality interference. Specifically, by integrating with the auxiliary detection heads in the RSC module, the MD module ensures that each unimodal branch backbone network receives only the optimization gradient signals corresponding to its designated representation space. This achieves modality-decoupled optimization of the backbone networks and removes the interference caused by $e^ {-(W_{k}^{m_i} f_i^{m_i})}$. 

The detailed procedure of the MD module can be expressed as follows:

\begin{equation}
\label{deqn_ex23}
\begin{cases}
\frac{\partial \, \mathrm{MD}^{(j)}}{\partial \, \mathrm{MD}_{(i)}}  \;=\; 0,   i \ \neq j 

\\[2ex]

\frac{\partial \, \mathrm{MD}^{(j)}}{\partial \, \mathrm{MD}_{(i)}}  \;=\; 1,   i \ = j
\end{cases}
\end{equation}
Here, $\mathrm{MD}_{(i)}$ denotes the $i$-th input branch of the MD module during the network forward computation, and $\mathrm{MD}^{(j)}$ denotes the $j$-th output branch of the MD module. Specifically, $i=0,1$ correspond to the feature map inputs of the two modalities, while $j=0,1,2$ correspond to the two auxiliary detection heads $Aux^{m_1}$, $Aux^{m_2}$, and the fusion module along with its detection head after passing through the MD module. Through the gradient mapping mechanism of the modality decoupling module, the gradients backpropagated from the newly introduced independent auxiliary detection heads are individually mapped to the corresponding feature-extraction backbones. This enforces a representation learning constraint consistent with unimodal learning, while simultaneously eliminating gradient interference from the fusion module and unrelated detection heads via gradient masking. Consequently, decoupled training of the modality-specific backbones is achieved, ensuring that each backbone can independently optimize towards its optimal representation. Furthermore, modality decoupling mitigates the imbalanced optimization between strong and weak modalities, preventing optimization conflicts arising from competitive learning among modalities.

In summary, under the proposed architecture, the multimodal object detection framework adopts the following paradigm: the loss generated by the fusion module and its detection head constrain the representation learning of the fusion subnetwork, whereas the representation learning of the unimodal branch backbones used for modality-specific feature extraction is supervised by the additional independent auxiliary detection heads. This design effectively prevents modality optimization conflicts and imbalanced learning among unimodal branches caused by modality coupling, thereby enhancing the generalization capability and robustness of the model.

\section{Experiments and Analysis}
\subsection{Public Datasets and Evaluation Metrics}
\subsubsection{FLIR} The FLIR\cite{ref61} dataset is a challenging multimodal object detection benchmark encompassing a variety of scenarios, including dark environments, heavy fog, smoke, adverse weather conditions, and glare. It provides both visible and infrared multimodal images, with the primary aim of encouraging researchers to exploit infrared or LiDAR features to complement and enhance visible-spectrum image representations. Although the dataset contains over 20,000 images, some are misaligned; therefore, following\cite{ref13}, we adopt a filtered, aligned subset. This aligned version comprises 5,142 image pairs, partitioned according to the official split, with 4,129 pairs allocated for training and 1,013 pairs for validation. Consistent with prior studies, categories with very few instances, such as dogs, are excluded, and only the people, car, and bicycle classes are retained for the experiments conducted in this work.

\subsubsection{LLVIP} The LLVIP\cite{ref3} dataset is a manually aligned visible-infrared multimodal object detection dataset. It is specifically designed for low-light detection tasks, and consequently, the majority of images in this dataset exhibit low illumination and predominantly dark environmental conditions. The dataset comprises 15,488 image pairs, of which 12,025 pairs are allocated for training and 3,463 pairs for testing.

\begin{table*}
\centering
\caption{Performance Comparison of Different Methods on FLIR Dataset.}
\label{tab1}
\begin{tabular}{c| c | c | c | c | c | c|  c }
\toprule
\multirow{2}{*}{Modality} & \multirow{2}{*}{Methods} & \multicolumn{3}{c|}{AP50(\%)} & \multirow{2}{*}{mean AP50(\%)} & \multirow{2}{*}{mean AP75(\%)} & \multirow{2}{*}{mean AP50-95(\%)} \\
\cline{3-5}
 & & Person & Car & Bicycle & & & \\
\midrule
VIS & Faster R-CNN & - & -& - & 65.0 & 22.8 & 30.2 \\
VIS & YOLOV5   & - & - & -& 67.8 & 25.9 &  31.8\\
VIS & SSD      & - &- & -& 52.2  & -    &  21.8\\
\midrule
IR & Faster R-CNN & - & - & - & 74.4& 32.5 & 37.6 \\
IR & YOLOV5   & - & - & - & 73.9 & 35.7 & 39.5 \\
IR & SSD      & - & - & - & 65.5 & 32.4 & 29.6 \\
\midrule
VIS + IR & CFT \cite{ref8}       &  80.4 & 90.2 & 61.4 & 72.9 & 30.9 & 37.3 \\
VIS + IR & ICAfusion \cite{ref21} &  81.6 & 89   & 66.9 & 79.2 & 36.9 &  40.8\\
VIS + IR & UniRGB-IR \cite{ref64} &  -&  -&    - & 81.4          & 40.2 &  44.1\\
VIS + IR & EI$^2$Det \cite{ref13} &  84.9 & 89.4 & 66.3 & 80.2 & -    & - \\
VIS + IR & LIF \cite{ref5}       &  -& -&- & -& -& 45.2 \\
VIS + IR & RCS-MD (ours) & 85.4 & 91.3 & 67.7 & \textbf{81.5} & \textbf{47.2} & \textbf{47.8} \\
\bottomrule
\end{tabular}
\end{table*}
\subsubsection{M3FD} The M3FD\cite{ref62} dataset is a multimodal object detection dataset captured using a binocular optical camera and a binocular infrared sensor, with the aligned visible and infrared images having a resolution of 1024 $\times$ 768. The dataset encompasses multiple scenarios, including daytime, cloudy, and nighttime conditions, and contains a total of 4,200 VIS-IR image pairs with six object categories: person, car, bus, motorcycle, truck, and Lamp, totaling 33,603 instances. However, the dataset does not provide predefined training and testing splits. For fairness, this work adopts the same training-testing partition as reported in this study\cite{ref13}, and all subsequent comparative evaluations are conducted based on the data splits from this reference.

\subsubsection{MFAD} The MFAD\cite{ref13} dataset contains visible-infrared image pairs captured under diverse weather conditions and encompasses a wide range of scene categories, including roads, tunnels, overpasses, parks, and parking lots. It covers images acquired under various visibility conditions, such as sunny, cloudy, foggy, and rainy weather, providing richer feature diversity. The dataset consists of 12,370 VIS-IR image pairs and provides an official training and testing split, with 9,879 pairs allocated for training and the remaining 2,473 pairs for testing. Six object categories are annotated in the dataset: car, bus, truck, pedestrian, EbikeRider (electric bicycle riders), and cyclist.

\subsubsection{Mean Average Precision(mAP)}mAP is one of the most representative metrics for evaluating model performance in object detection. It assesses model capability based on classification accuracy of predicted results and the Intersection over Union (IOU) between predicted bounding boxes and ground truth boxes. This paper employs $mAP_{50}$ and $mAP_{50-95}$ for evaluation, where $mAP_{50}$ denotes the average AP across all classes at an IOU threshold of 0.5, while $mAP_{50-95}$ denotes the average mAP calculated at IOU thresholds ranging from 0.5 to 0.95 in 0.05 increments. Higher mAP values indicate superior model performance.

\begin{table*}
\centering
\caption{Performance Comparison of Different Methods on LLVIP Dataset.}
\label{tab2}
\begin{tabular}{c| c | c | c|  c }
\toprule
Modality & Methods & mean AP50(\%) & mean AP75(\%) & mean AP50-95(\%) \\
\midrule
VIS & Faster R-CNN & 91.4 & 48.0 & 49.2  \\
VIS & SSD          & 82.6 & 31.8 & 39.8  \\
VIS & YOLOV7 X & 90.1 & 52.7 & 50.6  \\
VIS & YOLOV10 L & 87.1 & 53.9 & 50.5  \\
\midrule
IR & Faster R-CNN& 96.1 & 68.5 & 61.1  \\
IR & SSD         & 90.2 & 57.9 & 53.5  \\
IR & YOLOV7 X & 90.1 & 52 & 1  \\
IR & YOLOV10 L & 95.1 & 72.0 & 63.3  \\
\midrule
VIS + IR & CFT \cite{ref8}        & 97.5 & 72.9 &  63.6 \\
VIS + IR & ICAfusion \cite{ref21} & 97.4 & 70.9 &  62.7\\
VIS + IR & UniRGB-IR \cite{ref64} & 96.1 & 72.2 &  63.2 \\
VIS + IR & EI$^2$Det \cite{ref13} & 98 &  73.2 &   63.9\\
VIS + IR & LIF \cite{ref5} & - &  - &  67.9 \\ 
VIS + IR & RCS-MD (ours) & 97.7 & \textbf{81.7} & \textbf{69.5} \\
\bottomrule
\end{tabular}
\end{table*}

\begin{table*}
\centering
\caption{Performance Comparison of Different Methods on M3FD Dataset.}
\label{tab3}
\begin{tabular}{c| c | c | c | c | c | c|  c | c|  c|  c }
\toprule
\multirow{2}{*}{Modality} & \multirow{2}{*}{Methods} & \multicolumn{6}{c|}{AP50(\%)} & \multirow{2}{*}{mean AP50(\%)} & \multirow{2}{*}{mean AP75(\%)} & \multirow{2}{*}{mean AP50-95(\%)} \\
\cline{3-8}
 &               &  People & Car  & Bus  & Motorcycle & Lamp  &  Truck &      &   &   \\

\midrule
VIS & YOLO\_v5 L & 72.2    & 90.5 & 91.2 & 73.1       & 82.9  &  86.1  & 82.7 & 54.9  & 52.5  \\
VIS & YOLO\_v7 X & 75.6    & 91.2 & 92.0 & 75.3       & 84.7  &  88.4  & 84.9 & 56.0  & 53.2  \\
VIS & YOLO\_X L  & 70.3    & 88.6 & 89.0 & 71.2       & 79.1  &  82.9  & 80.3 & 53.5  & 51.3  \\
\midrule
IR & YOLO\_v5 L  & 81.3    & 87.2 & 86.8 & 73.4       & 70.5  &  82.5  & 80.3 & 53.2 & 50.7  \\
IR & YOLO\_v7 X  & 83.9    & 88.7 & 87.1 & 74.0       & 61.2  &  86.7  & 80.5 & 50.7 & 49.9  \\
IR & YOLO\_X L   & 79.6    & 85.3 & 84.9 & 67.8       & 60.1  &  79.5  & 74.6 & 51.3 & 49.1  \\
\midrule
VIS + IR & CFT \cite{ref8} & 82.2    & 91.3 & 91.6 & 73.6       & 85.1  &  86.1  & 85.0 & 57.4  & 54.5  \\
VIS + IR & 
       ICAfusion \cite{ref21} & 83.0    & 91.0 & 92.3 & 76.5       & 85.0  &  88.9  & 85.1 & 56.4  & 53.5  \\
VIS + IR & 
       MMI-Det \cite{ref11}   & 80.6    & 90.7 & 89.5 & 70.7       & 83.3  &  85.9  & 83.5 & 54.6  & 51.9  \\
VIS + IR &
       EI$^2$Det \cite{ref13} & 82.8    & 91.4 & 91.6 & 77.2       & 84.7  &  89.3  & 86.2 & 59.2  & 55.5  \\
VIS + IR & RCS-MD (ours)   & \textbf{85.2} & \textbf{92.4} & \textbf{91.7} & 73.2  & 82.3 & 86.2 & 85.2 & \textbf{64.7} & \textbf{59.5} \\
\bottomrule
\end{tabular}
\end{table*}

\subsection{Implementation Details}
This paper conducts optimization using the SGD optimizer, with a default initial learning rate of $1 \times e^{-2}$. The momentum is set to 0.937, the learning rate gradually decays to $1 \times e^{-6}$, and the weight decay is set to $1.0 \times e^{-5}$. Consistent with\cite{ref13}, the training image input size is set to  $640 \times 640$, and the testing image input size is also set to $640 \times 640$. Data augmentation techniques include mosaic data augmentation and random flipping. To ensure fair comparison with current SOTA models, minor parameter adjustments were made across datasets. These adjustments will be detailed in the results comparison section. Unless otherwise specified, the implementation details outlined above are used by default.The implementation details of each module in the network are same as that for the YOLOv8 network.
\subsection{Comparison with the current state-of-the-art Methods}
\subsubsection{Results on FLIR} On the FLIR dataset, we compare the proposed method with five representative prior approaches as well as the latest state-of-the-art (SOTA) models, including both multimodal and unimodal detectors, as summarized in Table \ref{tab1}. The proposed RSC-MD achieves SOTA performance on FLIR, obtaining 47.8\% $mAP_{50-95}$ and 81.5\% $mAP_{50}$, which represents an improvement of approximately 2.6\% over the most recent SOTA method\cite{ref5}. RSC-MD delivers the best AP across all categories (person, car, and bicycle). Specifically, using a model configured with a parameter scale comparable to YOLOv8-M, RSC-MD attains an $AP_{50}$ of 85.4\% for the person class, an $AP_{50}$ of 91.3\% for the car class, and an $AP_{50}$ of 67.7\% for the bicycle class. Although our model maintains the same parameter scale as YOLOv8-M, it surpasses transformer-based detectors with substantially larger parameter capacities. When evaluating the stricter $mean AP_{75}$ metric, RSC-MD exhibits an even more pronounced performance advantage over all comparison methods.

\subsubsection{Results on LLVIP} On the LLVIP dataset, Table \ref{tab2} presents a performance comparison between the proposed RSC-MD method and several recent multimodal approaches, with the best results highlighted in bold. RSC-MD likewise achieves state-of-the-art performance on LLVIP, surpassing the previous SOTA by approximately 1.6\% and reaching 69.5\% in terms of $mAP_{50-95}$. These results consistently exceed those of existing multimodal learning methods, confirming that the optimization deficiencies identified in this work indeed suppress model performance. In terms of $AP_{50}$, RSC-MD attains 97.7\%, which is slightly lower than the result reported in\cite{ref13}. This minor discrepancy arises because the study\cite{ref13} employs a more complex modality fusion mechanism, whereas the proposed method adopts only a naive element-wise addition for feature fusion. Nonetheless, the difference between the two models in $AP_{50}$ remains small.

\subsubsection{Results on M3FD}Since the M3FD dataset does not provide an official division of the training and testing sets, this work adopts the same experimental setting as the latest study\cite{ref13} to enable a fair comparison. In that study, 2,100 image pairs out of the 4,200 pairs were used for training, and the remaining 2,100 pairs were used for validation. As shown in Table \ref{tab3}, the proposed RSC-MD method achieves state-of-the-art performance in terms of $mAP_{50-95}$ metric,attaining a 4\% improvement over the best existing model under the same setting. For the $mAP_{50}$ metric, the proposed method obtains the highest performance for the person, car, and bus categories. Although it does not achieve the best performance for the motorcycle and lamp categories, careful analysis indicates that this is caused by two reasons. First, an analysis of the M3FD dataset reveals that its images were captured using an in-vehicle camera under motion, resulting in substantial motion blur and other negative degradations that severely affect image features, which is unfavorable for multimodal fusion-based detection. Second, the modality fusion method adopted in this work is naive addition, which is the most basic and primitive fusion strategy. Therefore, it may not fully preserve effective features from all modalities, and the noise contained in the modality-specific features may be propagated to the fused representations. 

However, these two points are not the focus of this study. The proposed RSC-MD method can be combined with any modality fusion strategy for detection, and thus we do not discuss these aspects in detail.

\subsubsection{Results on MFAD} Table \ref{tab4} presents the detection results on the MFAD dataset. On this dataset, the proposed RSC-MD method also achieves state-of-the-art performance. While obtaining the best $mAP_{50-95}$, the method does not reach the highest performance for a few individual categories. Specifically, RSC-MD yields a 3.7\% improvement in $mAP_{50-95}$, achieving a score of 57\%. It attains the best $AP_{50}$ results for large-object categories such as car, bus, and truck. Although it ranks second for the pedestrian and bicycle categories, the gap compared with the best result is only 0.1\%. This slight discrepancy is consistent with the reasons analyzed for the previous datasets, and thus is not reiterated here.

\begin{figure}[!t]
\centering
\includegraphics[width=3.4in]{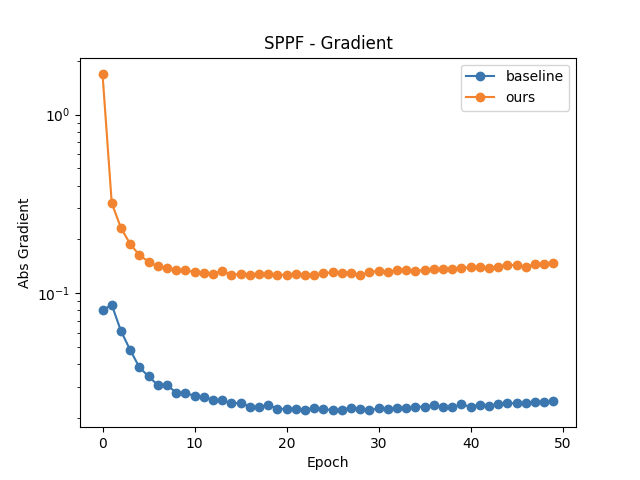}

\caption{Visual comparison of gradients in SPPF layers of visible modality.}
\label{gard_vis}
\end{figure}

\begin{figure}[!t]
\centering
\includegraphics[width=3.4in]{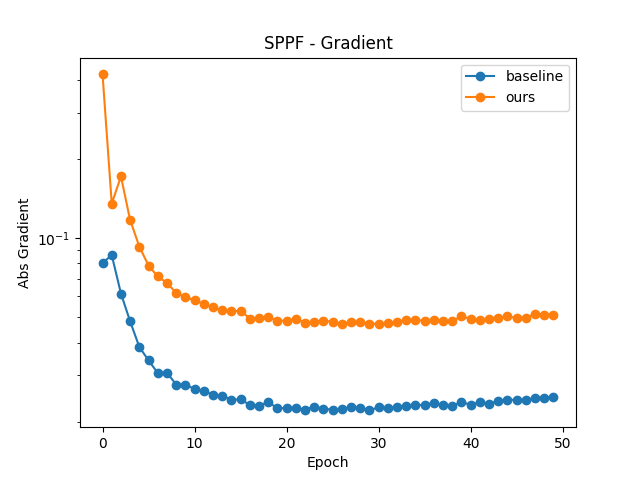}

\caption{Visual comparison of gradients in SPPF layers of infrared modality.}
\label{gard_ir}
\end{figure}
\begin{table*}
\centering
\caption{Performance Comparison of Different Methods on MFAD Dataset.}
\label{tab4}
\begin{tabular}{c| c | c | c | c | c | c|  c | c|  c|  c }
\toprule
\multirow{2}{*}{Modality} & \multirow{2}{*}{Methods} & \multicolumn{6}{c|}{AP50(\%)} & \multirow{2}{*}{mean AP50(\%)} & \multirow{2}{*}{mean AP75(\%)} & \multirow{2}{*}{mean AP50-95(\%)} \\
\cline{3-8}
 & & Car & BUs & Tru &Ped & Ebi& Cyc& & & \\
\midrule
VIS & YOLOV7\_X  & 89.5 & 84.9 & 82.6 & 65.8 & 71.9 & 42.4 & 72.9 & 52.7 & 48.8  \\
VIS & YOLOV10\_L & 88.1 & 85.1 & 80.4 & 61.8 & 66.8 & 44.1 & 71.1 & 53.3 & 48.9 \\
VIS & YOLOX\_L   & 80.3 & 73.5 & 75.2 & 57.1 & 65.3 & 41.5 & 68.6 & 52.2 & 46.8 \\
\midrule
IR & YOLOV7\_X   & 81.7 & 80.6 & 72.5 & 68.1 & 62.1 & 34.0 & 66.5 & 42.4 & 40.6 \\
IR & YOLOV10\_L  & 81.1 & 79.9 & 72.4 & 63.3 & 59.1 & 38.4 & 65.7 & 43.9 & 41.8 \\
IR & YOLOX\_L    & 79.6 & 76.4 & 71.3 & 61.1 & 56.6 & 37.6 & 63.2 & 41.5 & 39.2 \\
\midrule
VIS + IR & CFT \cite{ref8}  & 89.5 & 89.0 & 84.5 & 75.9 & 75.3 & 51.0 & 77.8 & 56.8 & 52.5 \\
VIS + IR & ICAfusion \cite{ref21} & 89.2 & 88.6 & 86.1 & 70.9 & 75.3 & 50.2 & 77.6 & 57.2 & 52.7 \\
VIS + IR & TINet \cite{ref65} & 84.1 & 84.3 & 77.2 & 67.6 & 63.8 & 42.6 & 69.1 & 45.5 & 43.6 \\
VIS + IR & MMI-Det \cite{ref11}& 89.6 & 88.2 & 86.0 & 76.9 & 75.2 & 45.2 & 76.9 & 55.9 & 51.4 \\
VIS + IR & EI$^2$Det \cite{ref13} & 89.8 & 88.8 & 85.7 & 78.6& 75.5& 53.3& 79.0 & 58.0 & 53.3 \\
VIS + IR & RCS-MD (ours) & \textbf{90.9} & \textbf{90.6} & \textbf{86.9} & 78.2 & \textbf{76.9} & 53.2 & \textbf{79.4} & \textbf{62.0} &  \textbf{57.0} \\
\bottomrule
\end{tabular}
\end{table*}

\begin{table}
\centering
\caption{Ablation Study on FLIR Dataset.}
\label{tab5}
\begin{tabular}{ c  c | c | c | c }
\toprule
RSC & MD & {mean AP50(\%)} & mean AP75(\%) & mean AP50-95(\%) \\
\midrule
- & - & 75.5 & 42.1 & 43.5 \\
\checkmark & - &  79.9 & 45  & 45.9   \\
\checkmark& \checkmark &  \textbf{81.5} & \textbf{47.2} & \textbf{47.8}\\
\bottomrule
\end{tabular}
\end{table}

Although the RSC-MD method incurs a slight increase in computational cost during training due to the introduction of two additional detection heads, its overall parameter count remains substantially smaller than that of complex distillation-based approaches and Transformer-based methods with large model capacities. Moreover, during inference, the auxiliary detection heads can be discarded, and only the fusion detection head is retained, resulting in a model whose parameter size is identical to that of the naive-addition baseline and introducing no additional parameters. Compared with complex distillation strategies and parameter-intensive Transformer architectures, the proposed method therefore achieves an efficient and lightweight parameter optimization.
\begin{table}
\centering
\caption{Ablation Study on LLVIP Dataset.}
\label{tab6}
\begin{tabular}{ c  c | c | c | c }
\toprule
RSC & MD & {mean AP50(\%)} & mean AP75(\%) & mean AP50-95(\%) \\
\midrule
- & - & 96.9 & 75.8 & 65.9 \\
\checkmark & - & 97.3 & 77.9 & 67.3 \\
\checkmark& \checkmark & \textbf{97.7} & \textbf{81.7} & \textbf{69.5} \\
\bottomrule
\end{tabular}
\end{table}
\subsection{Ablation Studies}
The ablation studies are conducted on the FLIR and LLVIP datasets. We first examine the individual contributions of the two modules and subsequently investigate the hyperparameters on the FLIR dataset. Consistent with widely adopted practice in prior studies, this work employs naive addition as the baseline, whose architecture is illustrated in Figure \ref{naiveAdd}. 

The RSC and MD modules are then progressively integrated into the baseline to assess the respective contributions of each component. It is noted that the modality decoupling module must be used in conjunction with the RSC module; when applied in isolation, the modality decoupling module yields zero gradients for the backbone networks and thus renders them untrainable. Therefore, the case of using the modality decoupling module alone is not considered. For clarity of presentation, the model using naive addition is referred to as the $baseline$ model. When the RSC module is added to the baseline, the resulting model is denoted as $baseline_{RSC}$. Finally, when the MD module is incorporated into the $baseline_{RSC}$ model, it is referred to as the $RSCMD$ model.

\subsubsection{Ablation Study on the FLIR Dataset} To evaluate the effectiveness of the RSC-MD method, the two modules were incrementally incorporated into the baseline model to verify their impact on model performance. As shown in Table \ref{tab5}, the $baseline_{RSC}$ model, which includes the RSC module, achieves an improvement of 4.4\% in $mAP_{50}$, 2.9\% in $mAP_{75}$, and 2.4\% in $mAP_{50-95}$ compared with the $baseline$ model. The $RSCMD$ model, relative to the $baseline$ model, attains increases of 6\% in $mAP_{50}$, 5.1\% in $mAP_{75}$, and 4.3\% in $mAP_{50-95}$. Compared with the $baseline_{RSC}$ model, $RSCMD$ exhibits gains of 1.6\% in $mAP_{50}$, 1.8\% in $mAP_{75}$, and 1.9\% in $mAP_{50-95}$. These results indicate that the sequential addition of the RSC and MD modules progressively enhances the model performance, thereby validating the effectiveness of the proposed method.

\subsubsection{Ablation Study on the LLVIP Dataset} Similar to the experiments conducted on the FLIR dataset, the two modules were incrementally incorporated into the baseline model to evaluate their effects. As shown in Table \Ref{tab6}, the $baseline_{RSC}$ model on the LLVIP dataset achieves a marginal improvement of 0.4\% in $mAP_{50}$ compared with the baseline model, while demonstrating increases of 2.1\% in $mAP_{75}$ and 1.4\% in $mAP_{50-95}$. The $RSCMD$ model, relative to the baseline model, attains improvements of 0.8\% in $mAP_{50}$, 5.9\% in $mAP_{75}$, and 3.6\% in $mAP_{50-95}$. Compared with the $baseline_{RSC}$ model, $RSCMD$ further improves performance by 0.4\% in $mAP_{50}$, 3.8\% in $mAP_{75}$, and 2.2\% in $mAP_{50-95}$. These results demonstrate that the sequential addition of the RSC and MD modules consistently enhances model performance on this dataset, thereby validating the effectiveness of the proposed approach.

\subsubsection{Ablation Study of Hyperparameters}This paper investigates the impact of different parameters on model performance within the FLIR dataset, as illustrated in the Table \ref{tab7}. 
This study finds that, unlike classification tasks where increasing the loss weight typically improves performance, excessively large hyperparameters in dense prediction tasks such as object detection can cause the optimization direction to deviate from the optimum, thereby degrading model performance.
Analysis reveals that unlike the imbalance observed in classification tasks, the visible and infrared images studied here exhibit relatively smaller modal differences compared to typical classification tasks such as image and speech. The two modalities are comparatively similar. Consequently, significant differences exist in both data distribution and scale compared to classification tasks. Therefore, in the field of multimodal detection, it is appropriate to seek the optimal model performance through small adjustments of hyperparameters.

\subsection{Regression verification of theoretical analysis} Through theoretical analysis, two optimization deficiencies are identified in this study: (1) the backbone of a unimodal branch within a multimodal detection model experiences substantial gradient suppression due to optimization conflicts, resulting in under-optimization of the unimodal branch; and (2) the weak modality is subjected to greater gradient suppression than the strong modality, leading to imbalanced optimization across unimodal branches. 

In this section, this study experimentally validates the two defects to confirm the consistency between the experimental results and the theoretical derivations.

Regarding \textbf{Optimization Deficiency (1)}, this paper conducted a gradient visualization study at the SPPF layer, as shown in Figures \ref{gard_vis} and \ref{gard_ir}. It can be observed from the figures that the gradient magnitude at the SPPF layer of the unimodal branch backbone, when employing the proposed method, is several times larger than that obtained using the baseline method with naive addition fusion. This observation is consistent with the conclusion of \textbf{Optimization Deficiency (1)}, wherein the unimodal branch backbone suffers from gradient suppression, resulting in gradients that are significantly smaller than those of a single-modality backbone.

\begin{figure}[!t]
\centering
\includegraphics[width=3.1in]{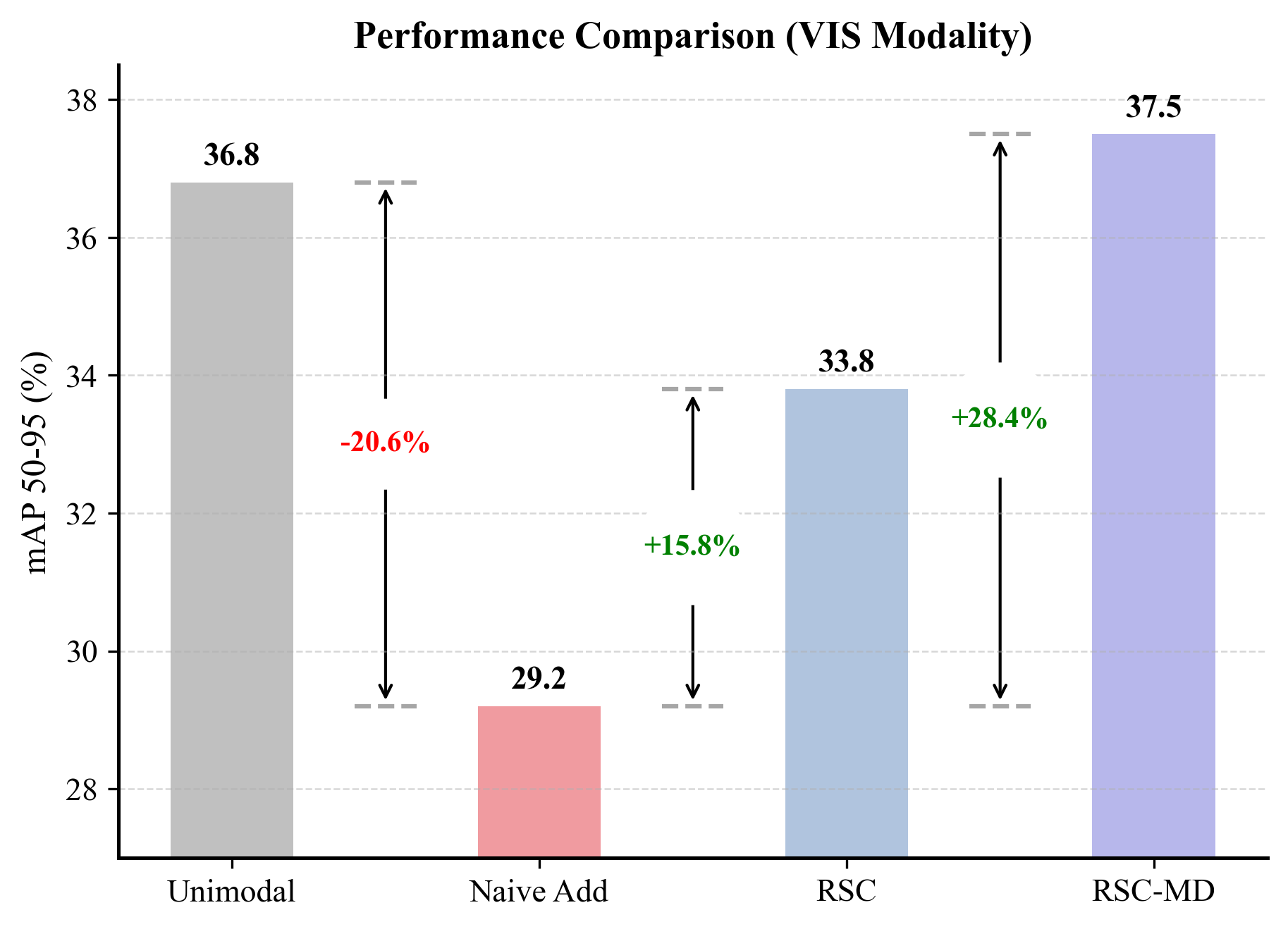}

\caption{Performance Comparison (VIS Modality).}
\label{rgb_perf}
\end{figure}

\begin{figure}[!t]
\centering
\includegraphics[width=3.1in]{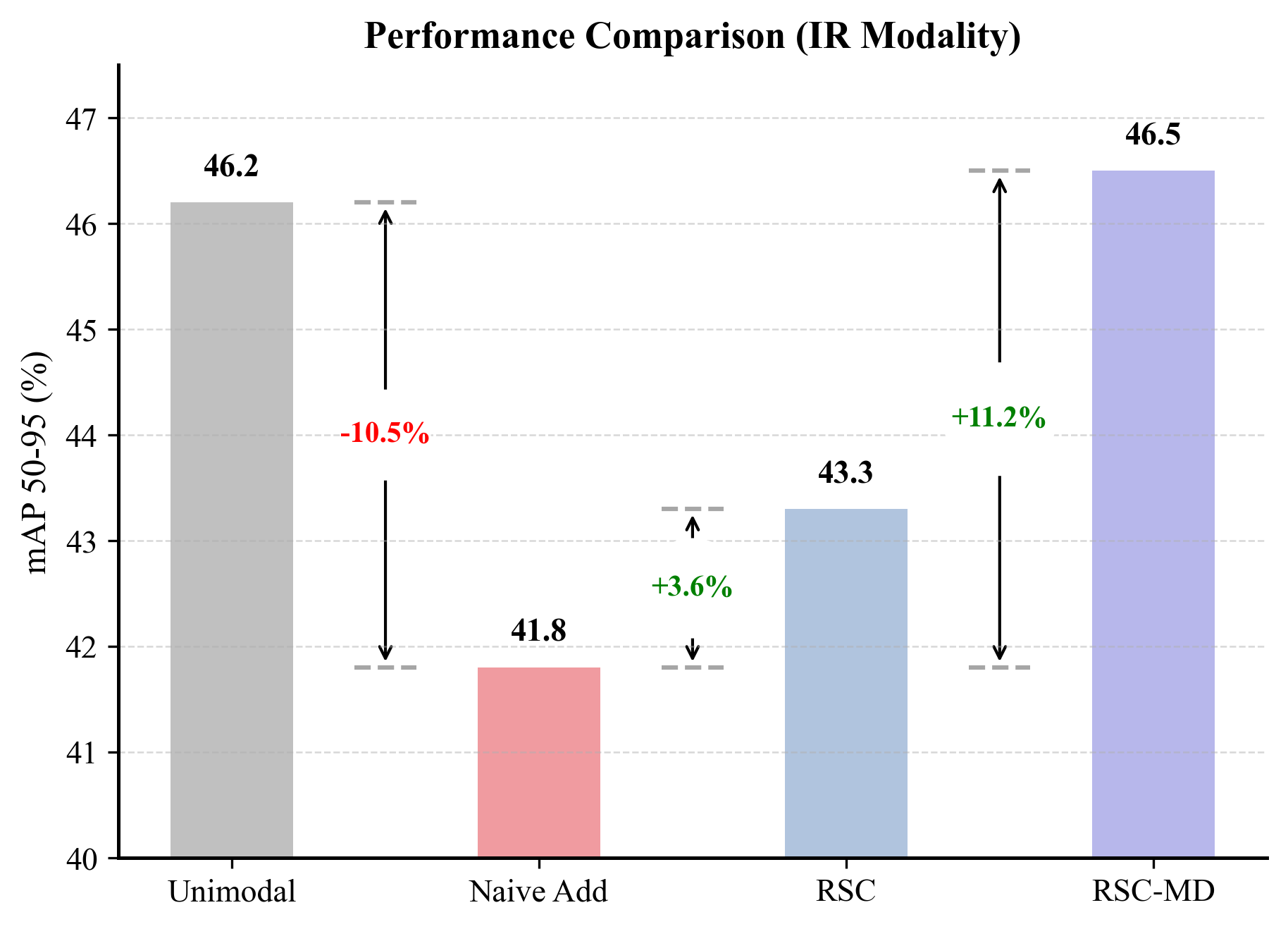}

\caption{Performance Comparison (IR Modality).}
\label{ir_perf}
\end{figure}

For \textbf{Optimization Deficiency (2)}, this study employs a linear evaluation consistent with the study\cite{ref5} and evaluates the performance of the corresponding unimodal branch. As shown in Figures \ref{rgb_perf} and \ref{ir_perf}, the VIS modality constitutes the weaker modality within the FLIR dataset. When naive addition is employed as the feature fusion method, its performance was reduced by 20.6\% due to gradient suppression. The IR modality experienced a performance reduction of 10.5\% due to gradient suppression. These results are consistent with our conclusions: the greater the gradient suppression experienced by the weaker modality, the more significant the impact on its performance.

Although certain studies\cite{ref5} have conducted linear evaluations from a unimodal perspective, they only reveal the under-optimization of unimodal branches based on the naive addition baseline, without validating or comparing the performance improvements of unimodal branches under the improved methods.
In contrast, this study not only identifies the root causes of unimodal learning deficiencies through theoretical analysis but also evaluates the post-improvement performance of both VIS and IR modalities.

\begin{table}
\centering
\caption{Ablation Study of Hyperparameters on FLIR Dataset.}
\label{tab7}
\begin{tabular}{ c c c | c | c | c }
\toprule
 {$\alpha$} & {$\beta$} & {$\gamma$} & {mean AP50(\%)} & mean AP75(\%) & mean AP50-95(\%) \\
\midrule
 1& 1 & 1 & 81.5 & 47.2 & 47.8 \\  
 1& 2 & 1 & 81.4 & 47.1 & 47.5 \\
 1& 2 & 2 & 81.7 & 46.0 & 47.4 \\
 1& 4 & 4 & 81.6 & 46.3 & 47.4 \\
 1& 5 & 5 & 80.2 & 46.6 & 47.1 \\
 2& 2 & 3 & 81.2 & 46.4 & 47.3 \\
 3& 1 & 1 & 81.9 & 45.9 & 47.3 \\
\bottomrule
\end{tabular}
\end{table}

As illustrated in Figures \ref{rgb_perf} and \ref{ir_perf}, the introduction of RSC and MD modules yields performance gains of 15.8\% and 28.4\% in the VIS modality, and 3.6\% and 11.2\% in the IR modality, relative to naive addition. This indicates that the VIS modality exhibits a larger improvement, consistent with the prior analysis that VIS experiences greater suppression under modal coupling and, consequently, benefits more substantially from the enhancements. Furthermore, the comparison between Figures 3 and 4 shows that the VIS modality in Figure 3 undergoes more severe gradient suppression than the IR modality in Figure 4, corroborating \textbf{Optimization Deficiency (2)}.

Additionally, Figures \ref{rgb_perf} and \ref{ir_perf} demonstrates that the RSC module alone cannot completely eliminate interference arising from modal coupling. Specifically, for the IR modality branch trained on the FLIR dataset, the incorporation of RSC to amplify suppressed gradients results in a 3.6\% performance improvement; however, due to interference from excessive gradients contributed under negative sample scenarios, the model still fails to reach the performance level obtained when trained exclusively on unimodal data. This observation aligns with the conclusions discussed in the RSC methodology.

\section{conclusion} This study provides a theoretical analysis of feature degradation in multimodal detection and reveals two key optimization deficiencies: under-optimization of unimodal branches and imbalanced modality learning. To address these issues, we propose a representation space constrained learning with modality decoupling approach, comprising an RSC module and an MD module, which mitigate gradient suppression from modality coupling and imbalance, enabling full optimization of each modality backbone. Extensive experiments on the FLIR, LLVIP, M3FD, and MFAD datasets demonstrate that the proposed RSC-MD consistently achieves state-of-the-art performance, improving $mAP_{50-95}$ by 2.6\% on FLIR and 1.6\% on LLVIP, while maintaining a lightweight parameter configuration. Importantly, the proposed method is independent of the specific modality fusion strategy, allowing seamless integration with different fusion approaches and offering broad applicability as well as substantial potential for further exploration in multimodal object detection.
\bibliographystyle{IEEEtran}
\bibliography{ref.bib}

\end{document}